# Machine Learning-Based Completions Sequencing for Well Performance Optimization

Liu, A.[b*], Sun, J.W.[b], Ngo, A.[b], Mabadeje, A.O.[a], and Hernandez-Mejia, J.L.[a]

a. Hildebrand Department of Petroleum & Geosystems Engineering, The University of Texas at Austin
b. College of Natural Sciences, The University of Texas at Austin
[*]Corresponding author E-mail: anjie.liu@utexas.edu

## Abstract

Establishing accurate field development parameters to optimize long-term oil production takes time and effort due to the complexity of oil well development, and the uncertainty in estimating long-term well production. Traditionally, oil and gas companies use simulation software that are inherently computationally expensive to forecast production. Thus, machine learning approaches are recently utilized in literature as an efficient alternative to optimize well developments by enhancing completion conditions. The primary goal of this project is to develop effective machine-learning models that can integrate the effects of multidimensional predictive variables (i.e., completion conditions) to predict 12-Month Cumulative Production accurately.

Three predictive regression machine learning models are implemented for predicting 12-month cumulative oil production: Random Forest, Gradient Boosting, and Long Short-Term Memory Models. All three models yielded cumulative production predictions with root mean squared error ($RMSE$) values ranging from 7.35 to 20.01 thousand barrels of oil. Although we hypothesized that all models would yield accurate predictions, the results indicated a crucial need for further refinement to create reliable and rational predictive tools in the subsurface. While this study did not produce optimal models for completion sequencing to maximize long-term production, we established that machine learning models alone are not self-sufficient for problems of this nature. Hence, there is potential for significant improvement, including comprehensive feature engineering, and a recommendation of exploring the use of hybrid or surrogate models (i.e., coupling physics reduced models and machine learning models), to ascertain significant contribution to the progress of completion sequencing workflows.

# 1. Introduction

The oil and gas industry has a collective net worth of 239 billion dollars, making up 8% of the gross domestic product of the United States (Carpenter 2023). Undoubtedly, the ability to generate such significant revenue and maintain consistent oil production is vital for meeting the needs of both consumers and industries. Achieving these goals and supporting sustainable growth within the sector requires companies to analyze past data on well performances, identify valuable attributes of highly productive wells, and optimize well development and completions. However, this process can be time-consuming and resource-intensive due to the complexity and uncertainty in estimating long-term production, requiring efficient simulation software (Ibrahim et al. 2022).

Over the years, machine learning models are touted as a panacea to solve such challenges by catering to the highly dimensional nature of subsurface datasets, which allows for quick analyses, key attributes identification of high-performing oil wells, and optimize completion conditions for long-term cumulative oil production. Luo et al. (2018) uses a neural network model to analyze how a combination of formation depth, thickness, and other predictive variables results in higher productivity in the Bakken play. Zhang et al. (2022) utilizes reinforcement learning neural networks, classical evolutionary algorithms, and surrogate-assisted optimization methods on two reservoir models and proposes a new method to attain higher net present value. Similarly, Teixeira et al. (2018) uses neural networks to analyze contributing variables and rank its contribution to high oil flow rate in the Namorado Field, located in Brazil.

The primary goal of this work is to compare the relative effectiveness of machine-learning models by integrating the effects of multidimensional predictive variables (completion conditions) to accurately predict the 12 Months Cumulative Production of the Bakken formation. To achieve this goal, three distinct machine-learning models: Random Forest, Gradient Boosting, and the Long Short-Term Memory neural network method are employed. While previous works apply individual machine learning algorithms, we compare multiple models to enhance the understanding on achieving an optimal completions' sequencing.

The methodology section contains a detailed description of data preprocessing, the three machine-learning models that are utilized, and the training and optimization framework for these models. Subsequently, the results section describes the overall effectiveness, accuracy, and precision of the trained machine learning models. This is done in an analytical and comparative manner for a subsurface dataset located in the Bakken. The discussion section includes insights into sources of potential biases and areas of potential improvements for the machine learning models when performing optimization of completions sequencing on a field scale. Lastly, the overall evaluations for and future directions are detailed in the conclusions section.

## 2. Background

This section describes the three machine learning models used, aimed at exploring its classification and prediction capability.

### 2.1 Random Forest Model

Random Forest provides a baseline of comparison of the models used in this study and is considerably efficient as compared to a single tree model (Sahin 2020). Segal et al. (2004) defines sample bootstrap as a method that generates a single decision tree from a sample in a dataset instead of training based on the entire data. The Random Forest algorithm, first formally introduced by Brieman (2001), generates a forest of decision trees randomly from the training set using bootstrap and averages the results from the trees. This is an effective tool of prediction with higher accuracy than traditional decision trees. The model has a training time complexity of $\mathcal{O}(n \log(n) \cdot d \cdot k)$, where $n$ is the number of training instances, $d$ is the dimensionality of the training data, and $k$ is the number of decision trees.

To build the Random Forest Model, draw a bootstrap sample from the training set. Then, build a decision tree from that bootstrap sample. To create the decision tree, recursively select some variables randomly, pick the best variable among them, and split the node into two daughter nodes until the minimum node size is reached (Hastie et al. 2009). Repeat the first step a desired number of times to generate an ensemble of decision trees. Since the outcome variable is numeric and continuous, a random forest regression algorithm is used in contrast to a random forest classification model. **Eq. 1** means to sum the predicting value from a certain number of decision trees $T_b$, averaging it by the total number to get the final prediction value of the random forest, where $\hat{f}_{rf}^B(x)$ is the average output of all random forest trees, each with an input of a bootstrapped dataset of the given dataset and $B$ is the number of bootstrapped samples, which is equal to the number of random forest trees (Hastie et al. 2009).

$$\hat{f}_{rf}^{B}(x) = \frac{1}{B}\sum_{b=1}^{B} T_b(x) \qquad (1)$$

## 2.2 Gradient Boosting Model

Gradient Boosting is the most optimum model among ensemble models (Sahin 2020). The Gradient Boosting algorithm generates sequential predictions based on sections of the dataset. This model is founded upon the mathematical concept involving a series of decision trees used to make predictions based on a set of input variables (Natekin, A. and Knoll 2013). The Big $\mathcal{O}$ complexity for this boosting training is derived from formular $\mathcal{O}(n \cdot m \cdot T \cdot d)$, where $n$ represents the number of data points, $m$ is the number of predictor features, $T$ is the number of trees, and $d$ is the maximum depth of the trees.

To train a gradient boosting model, a base value is established, which is set to the actual value from the real dataset as a foundation for subsequent calculations. Individual decision trees are then created to calculate residues for the prediction, which are multiplied by different learning rate values. The adjusted residual values generated by these sequential decision trees are then summed together and added to the base value to arrive at the final prediction. This approach allows the algorithm to iteratively improve the accuracy of its forecasts by minimizing the residual error at each step (Ma et al. 2022; Ibrahim et al. 2022). Equation (2) provides a mathematical representation of the gradient boosting model, where $Y_i$ represents the predicted output, $F(X_i)$ is the base model, $f_d(X_i)$ represents the $m^{th}$ decision tree with parameters $\theta$, and $\sum f_d(X_i)$ is the sum of all decision trees (Ibrahim et al, 2022).

$$Y_i = F(X_i) = \sum f_d(X_i), \ f_d \in F, i = 1, \ldots, n \qquad (2)$$

## 2.3 Long Short-Term Memory (LSTM) Model

The LSTM model addresses the limitations of the recurrent neural network (RNN), particularly the Vanishing Gradient Problem (Sherstinsky, 2020). Due to the updating of weight values during backpropagation, the gradient of the loss function can diminish significantly in RNN, leading to the Vanishing Gradient Problem. The LSTM model, depicted in **Fig. 1**, overcomes this issue by incorporating specialized layers to control information flow. The forget gate layer, using a sigmoid function, decides which information to allow through the gate. The input gate layer and hyperbolic tangent layer update the cell state, preserving relevant information. Finally, another sigmoid and hyperbolic tangent function produce the output and determine which aspects of the previous observation to forget (Sherstinsky, 2020).

The LSTM model's architecture thus ensures that relevant information is retained and properly propagated, effectively addressing the limitations of the RNN, and allowing for better training and learning in sequential data analysis.

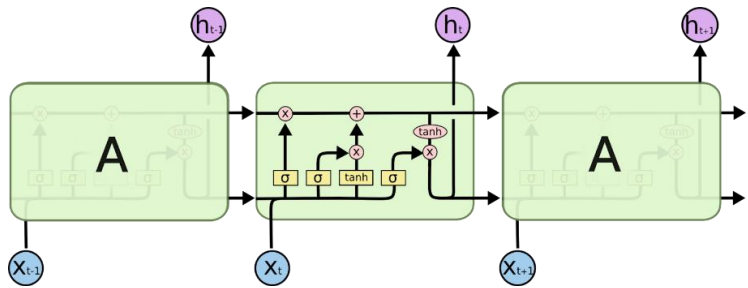

**Fig. 1**: the structure of a hidden layer of the LSTM model (McCullum, 2020).

## 3. Methodology

To evaluate the three models, a prediction workflow comprising of 7 main steps are adopted:

1. Preprocess the dataset and encoded predictor features using one-hot encoding method for non-ordinal categorical variables.
2. Split data into training and test sets to facilitate model predictions.
3. Train a machine learning model to make predictions of the response.
4. Evaluate the model using the Mean Squared Error ($MSE$), Correlation Coefficient ($R^2$), Root Mean Squared Error ($RMSE$) and Mean Absolute Error ($MAE$).

5. Perform hyperparameter tuning using the Bayesian Optimization package, Optuna, in Python.
6. Validate the tuned model on the test set using $MSE$, $R^2$, $RMSE$, and $MAE$ to evaluate the obtained predictions.
7. A model goodness check is conducted using confidence interval of predictions to account for epistemic uncertainty.

The first step of the proposed workflow involves removing predictor features with significant missing values or repeated information, encoding the categorical variables, filling in missing values using the Iterative Imputer from Scikit-Learn package, and separating variables with multifaceted information. Then, the resulting data is separated in a way such that 80% of it is the training set and 20% is the testing set. Next, a circular process is presented where hyperparameters are tuned using Optuna to train the model. Model evaluation is performed using the $MSE$, $R^2$, $RMSE$, and $MAE$ metrics, which is used to further refine and tune the hyperparameters until an optimal result maximizing each metric is obtained.

The tuned model is evaluated based on its epistemic uncertainty by comparing the distribution of MSE for a large number of realizations (100 and 500) while visualizing the cross-validated predictions. In machine learning model development, Scikit-learn was used for its wealth of algorithms and vast capabilities in model evaluation and validation, while TensorFlow was used to develop a neural network. Optuna facilitated hyperparameter optimization to enhance machine learning model performance coupled with external visualization libraries like Matplotlib and Seaborn.

# 4. Results
## 4.1 Data Description

Production wells from the Northwest region in North Dakota as shown in **Fig. 2b** is used as data.

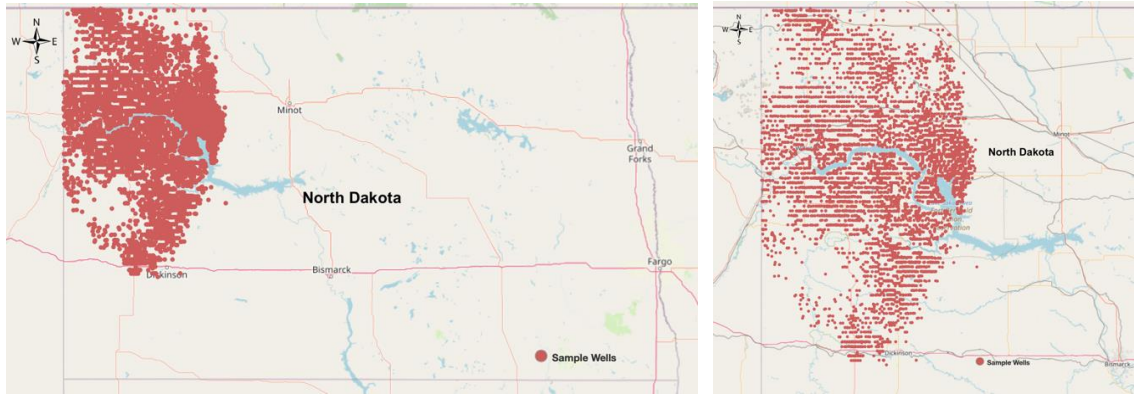

**Fig. 2.** (Left) Location of sample wells in the Northwestern corner of North Dakota, centering around Lake Sakakawea. (Right) A more zoomed-in look at the sample wells.

To address the missing predictor variable issue as shown in **Fig. 3b**, KNN-nearest neighbor imputation was implemented if missing values were below 25%; otherwise, the variable was dropped. Collinear and redundant variables as indicated in **Fig. 3a** were also removed, such as longitude and latitude, since *Township* and *Range* provide geographic information. While this approach may introduce some bias, it preserves sample size for analysis and machine learning model training. The Long Short Term Memory model requires an additional process of calculating the average number of barrels produced every month for the purpose of predicting the number of barrels produced the following month.

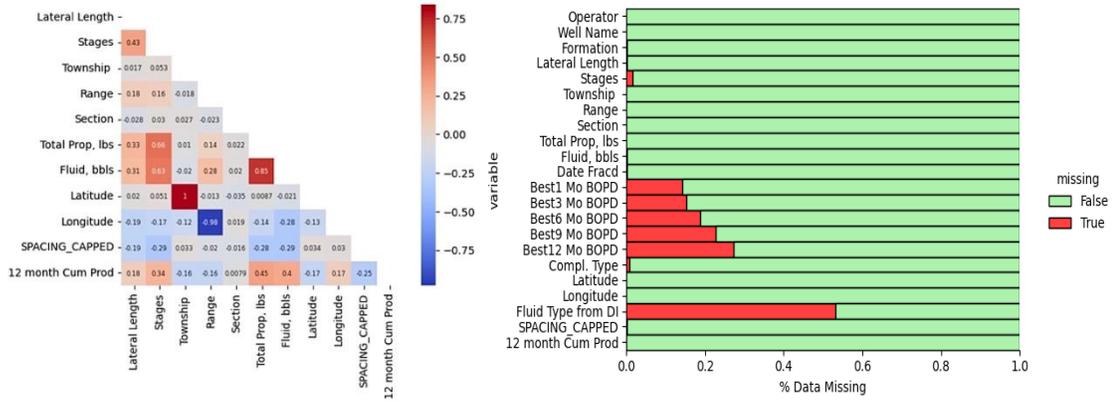

**Fig. 3.** (Left) Rank correlation coefficients of numerical features and (Right) Histogram showing the percentage of data values missing.

### 4.2 Random Forest Regression Model

The model was tuned utilizing the Optuna package for 100 trials. The resulting values of hyperparameters are n_estimater (200), max_depth (23), min_samples_split (3), and min_samples_leaf (2). The predicted model has a correlation coefficient of 0.86, with a *MSE* of 400.53, *RMSE* of 20.01, and *MAE* of 10.34. **Fig. 4** demonstrates that the predicted 12-month cumulative production lie relatively closely to the actual values, with one exception at above 450 thousand barrels.

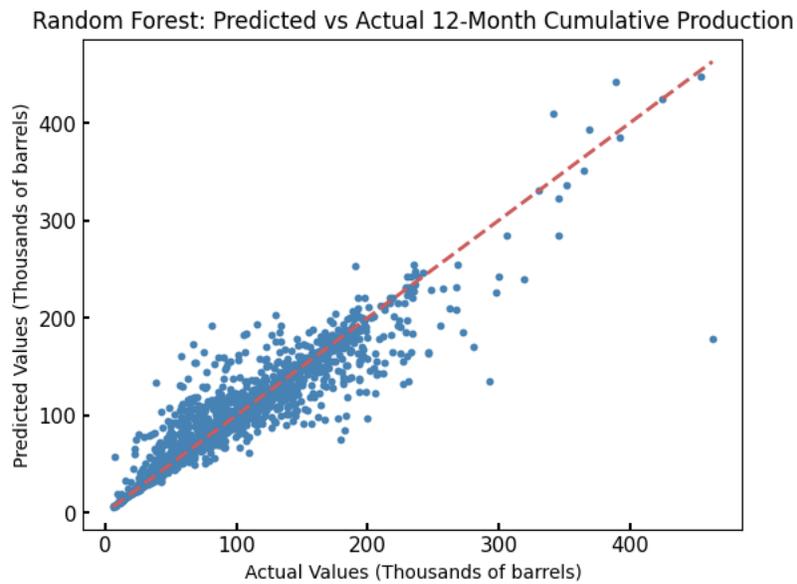

**Fig. 4.** Cross-validation of actual and predicted values of testing set for Random Forest Model

## 4.3 Gradient Boosting Model

The Gradient Boosting Machine Learning model was implemented with a sample split at the random state 121. Optuna was employed to optimize hyperparameters (100 trials, with goal of minimizing mean squared error, used 5 split K-fold cross validation), yielding the following optimal values: learning_rate (0.0219), max_depth (9), n_estimators (937), subsample (0.734), colsample_bytree (0.708), min_child_weight (5), gamma (0.275), and reg_alpha (0.998). The model achieved a correlation coefficient of 0.88, with a $MSE$ of 342.47, $RMSE$ of 18.51, and $MAE$ of 10.20. **Fig. 5** demonstrates that the predicted values lie relatively closely to the actual values.

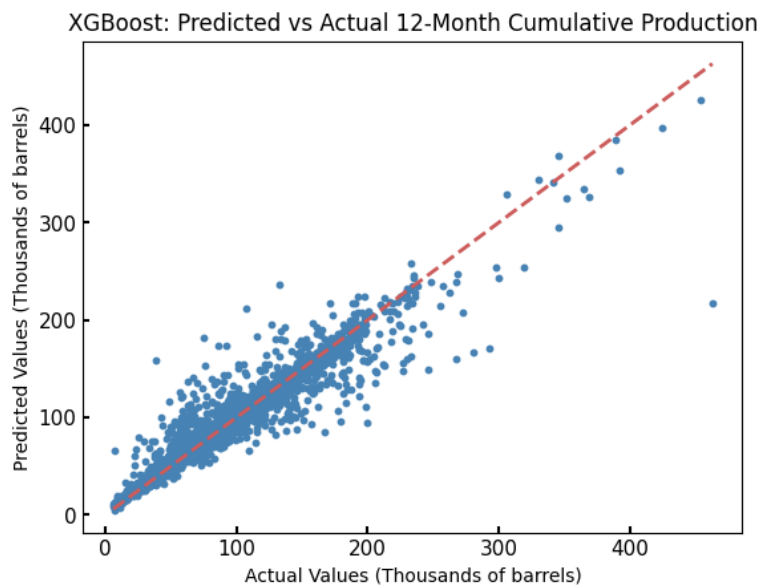

**Fig. 5.** Cross-validation of actual and predicted values of testing set for Gradient Boosting Model

## 4.4 LSTM Model

Due to the lack of information on cumulative production for consecutive days, the average of oil barrels produced per month was calculated and used as the response variable to predict the following month's cumulative production instead. The data was split into a training, validation, and testing set. The training set comprised 60% of the data while the validation and testing set contained 20% each. The LSTM model was then implemented with 2 layers. The first layer of the LSTM model contained 45 neurons while the second one contained 60 neurons with a dropout rate of 10% to avoid overfitting. The optimizer was chosen to be the Adam optimizer while the loss function was chosen to be $MSE$, which was also used to evaluate the training process. The hyperparameters were manually optimized and found to be a learning rate of 0.01, weight decay of 1e-6, epoch of 9, and batch size of 32. The *MSE* loss function for each epoch ranged from 0.23 to 0.53. The model produced an *MSE* value of 53.98, an *RMSE* of 7.35, an *MAE* of 5.59, and a correlation coefficient of 0.84. To evaluate the model, **Fig. 6** displays the output of the *MSE* loss function for the training and validation set for every epoch. After epoch 6, the curves for both the training and validation data briefly match, then they diverge slightly.

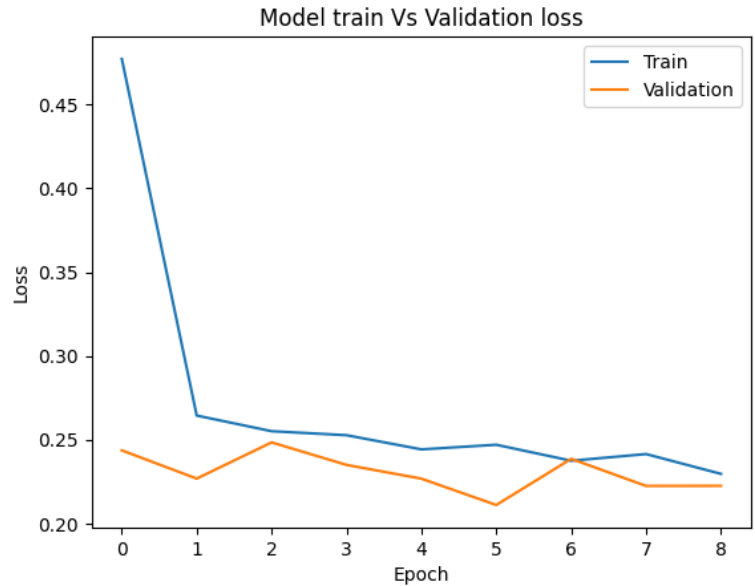

**Fig. 6.** Validation curve for LSTM model

**Fig. 7** provides a visual representation of how much the values predicted from the training set match the true values from the testing set. A source of error potentially lies in the possibly inadequate data cleaning process required for LSTM to learn efficiently.

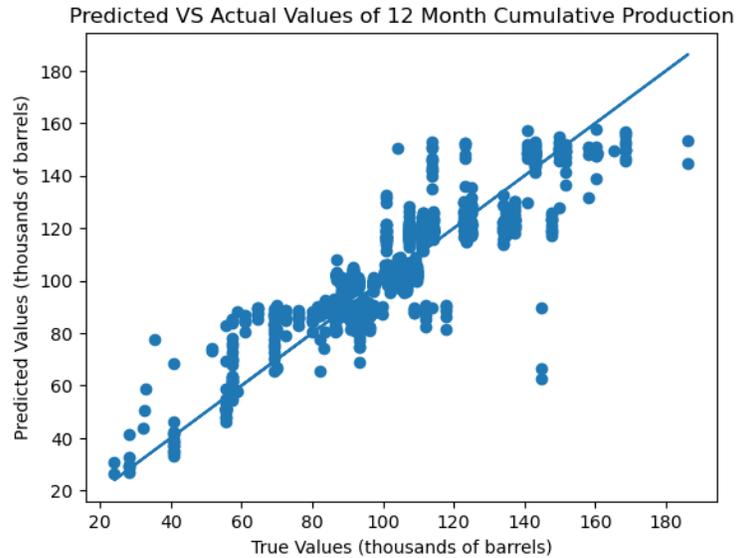

**Fig. 7.** Cross-validation of actual and predicted values for LSTM model.

## 5.  Discussion

### 5.1 Random Forest Regression Model

To obtain the epistemic uncertainty of the model, 100 and 500 realizations were run using the same hyperparameters and different random state values. As the sample size increases from 100 to 500, the 95% confidence interval in **Fig. 8** becomes tighter, with a lower bound of 19.965 and an upper bound of 20.252. This shows that the $RMSE$ value of this model is reliable with limited uncertainty.

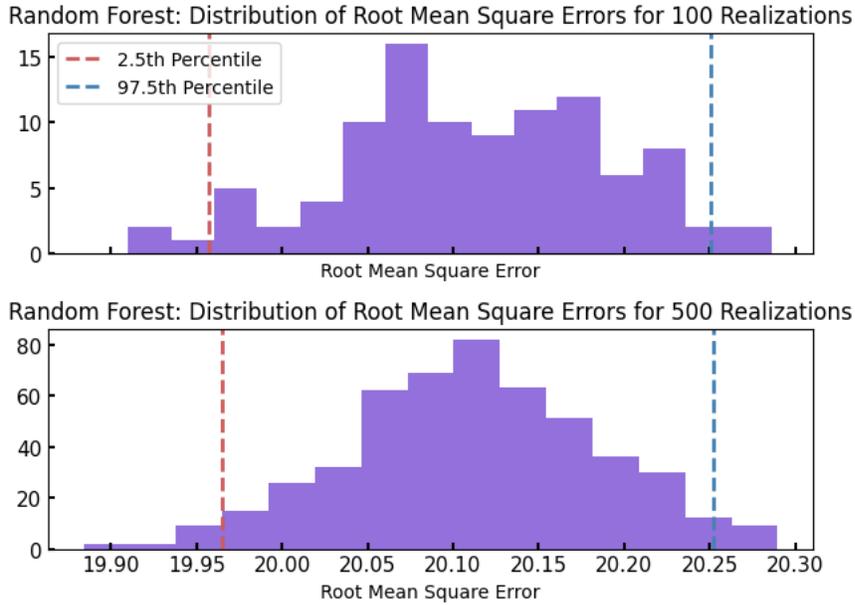

**Fig. 8**. Epistemic uncertainty associated with Random Forest Regression Model at 95% confidence intervals.

## 5.2 Gradient Boosting Model

The developed model yielded predictive regression results with a *MSE* of 342.47, evaluated against test data for accuracy. **Fig. 9** displays the model's uncertainty, derived from 100 and 500 bootstrap trials; the 500 bootstrap trials have a 95% confidence interval for *RMSE* between 11.189 and 18.510 thousand barrels of oil. Experimentation revealed that increasing test data proportions and decreasing training data proportions improved model accuracy; however, this could introduce bias due to insufficient test data for validation. Thus, an 80/20 train-test split was adopted to balance accuracy and validation. Further model enhancements could involve alternative feature selection methods, adjustments to data preprocessing techniques, or hyperparameter tuning. Potential improvements include modifying train-test split ratios, increasing Optuna hyperparameter tuning space with wider ranges, and additional feature engineering.

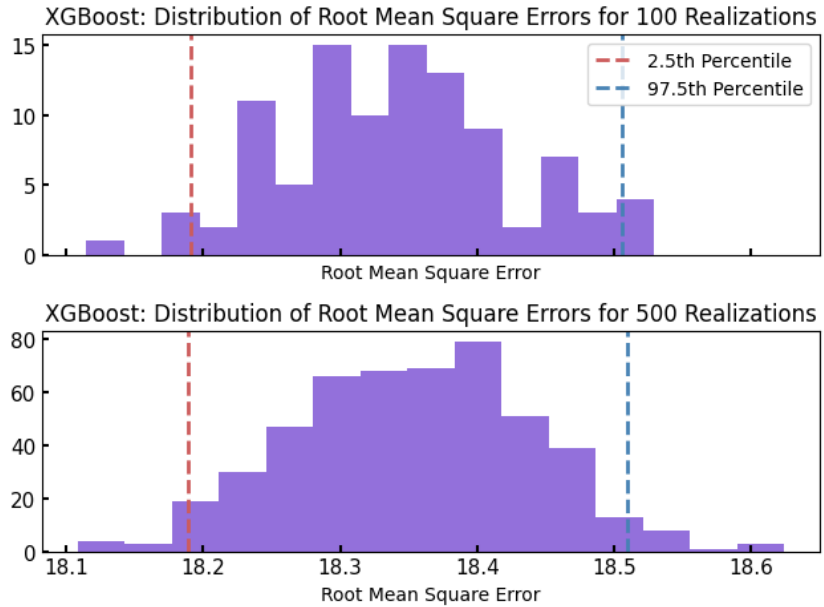

**Fig. 9**. Epistemic uncertainty associated with Gradient Boosting Model at 95% confidence intervals.

### 5.3 LSTM Model

**Fig. 10** displays the model's uncertainty after 40 trials, with a 95% confidence interval for *RMSE* between 7.37 and 10.09. A *MSE* of 53.98 indicates that the model does fairly good in forecasting values that are close to the actual values as compared to RF and XGBoost; the correlation coefficient of 0.84 confirms this result. To further validate the effectiveness of the LSTM model, **Fig. 5** shows that there is a small overfit as the *MSE* loss function for both the training and validation data diverge slightly after 6 epochs. However, the overfit remains within the 95% confidence interval for *RMSE*, which lies between 7.37 and 10.09 as shown in **Fig. 10**. This result suggests that the overfit is within an acceptable range and that the model performs fairly well. **Fig. 6** provides an illustration to further reinforce the observation that the number of oil barrels predicted from the training data closely matches the actual number of oil barrels recorded by the testing data.

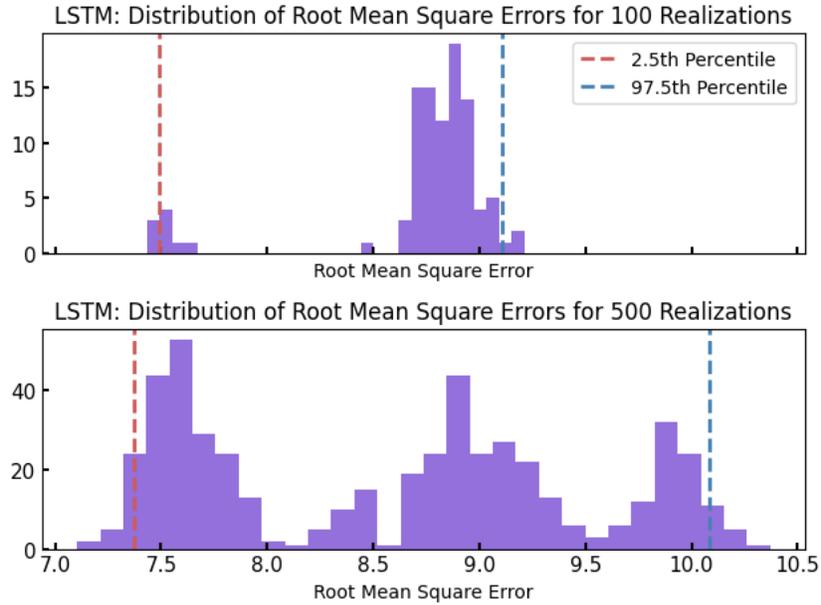

**Fig. 10**. Epistemic uncertainty associated with LSTM Model at 95% confidence intervals.

# 6. Conclusion

This study employs three machine learning models to assess various completion condition combinations, striving to optimize predicted cumulative oil production in wells through regression analysis. Although we hypothesized that all models would yield accurate predictions, our results indicated a need for further refinement to create reliable predictive tools. We recognize the potential for improvement, including comprehensive feature engineering and exploring alternative machine learning models.

Machine learning's implementation in industrial settings is an exciting and rapidly evolving domain, especially for the oil and gas industry, which can benefit immensely from the vast data accumulated over centuries of exploration. While this study did not produce optimal models for completion sequencing to maximize long-term production, our primary objective remains the development of advanced machine learning models to achieve optimal completion sequencing. We are dedicated to refining and enhancing our models, ensuring they significantly contribute to the progress of completion sequencing techniques in the oil and gas industry.

# Credit author statement

Anjie Liu: Data Curation, Conceptualization, Methodology, Codes and Data Analysis, Visualization, Writing – Original draft.
Jinglang W. Sun: Initial Approach Design, Codes and Data Analysis of XGBoost, Writing – Original draft
Anh Ngo: Codes and Data Analysis of LSTM, Writing – Original draft
Ademide O. Mabadeje: Supervision, Writing – Reviewing and Editing.
Jose L. Hernandez-Mejia: Supervision, Writing – Reviewing and Editing.

# Acknowledgment

The authors sincerely appreciate ConocoPhillips for providing the data, and supporting the College of Natural Sciences at the University of Texas at Austin on the Inventor Program.

# Data Availability

The data and well-documented workflow is on the corresponding author's GitHub Repository: https://github.com/anjieliu121/well_performance_optimization.